\documentclass[conference]{IEEEtran}
\IEEEoverridecommandlockouts

\usepackage{cite}
\usepackage{amsmath,amssymb,amsfonts}
\usepackage{algorithmic}
\usepackage{graphicx}
\usepackage{textcomp}
\usepackage{xcolor}
\usepackage{setspace}  
\usepackage{mathrsfs} 
\usepackage{multirow}
\usepackage[numbers]{natbib} 
\bibpunct{[}{]}{,\hspace{-0.05em}}{n}{}{,}  
\usepackage{booktabs}
\usepackage{caption}
\usepackage{hyperref} 

\usepackage[font=normalsize]{caption}  

\def\BibTeX{{\rm B\kern-.05em{\sc i\kern-.025em b}\kern-.08em
    T\kern-.1667em\lower.7ex\hbox{E}\kern-.125emX}}

\begin{document}

\title{KAN-HyperpointNet for Point Cloud Sequence-Based 3D Human Action Recognition
}

\author{
	\IEEEauthorblockN{
		Zhaoyu Chen\textsuperscript{1,2}\hspace{1em} 
		Xing Li\textsuperscript{1,*}\hspace{1em} 
		Qian Huang\textsuperscript{2}\hspace{1em} 
		Qiang Geng\textsuperscript{2}\hspace{1em} 
		Tianjin Yang\textsuperscript{2}\hspace{1em} 
		Shihao Han\textsuperscript{2}
	}
	\vspace{1em} 
	\IEEEauthorblockA{
		\textsuperscript{1}Nanjing Forestry Univeristy \hspace{3em} 
		\textsuperscript{2}HoHai University 
	}
}

\author{
    \IEEEauthorblockN{Zhaoyu Chen$^{1,2}$, Xing Li$^{2,*}$, Qian Huang$^{1}$, Qiang Geng$^{1}$, Tianjin Yang$^{1}$, Shihao Han$^{1}$}
    \IEEEauthorblockA{$^1$\textit{College of Computer Science and Software Engineering, Hohai University, Nanjing, China}}
    \IEEEauthorblockA{$^2$\textit{College of Information Science and Technology \& Artificial Intelligence, Nanjing Forestry University}}
    \IEEEauthorblockA{{\{chenzhaoyu,huangqian,qg,yangtianjin,hanshihao\}@hhu.edu.cn, lixing@njfu.edu.cn} 
\vspace{-0pt}
}}

\maketitle

\begin{spacing}{1.03} 
\begin{abstract}
Point cloud sequence-based 3D action recognition has achieved impressive performance and efficiency. However, existing point cloud sequence modeling methods cannot adequately balance the precision of limb micro-movements with the integrity of posture macro-structure, leading to the loss of crucial information cues in action inference. To overcome this limitation, we introduce D-Hyperpoint, a novel data type generated through a D-Hyperpoint Embedding module. D-Hyperpoint encapsulates both regional-momentary motion and global-static posture, effectively summarizing the unit human action at each moment. In addition, we present a D-Hyperpoint KANsMixer module, which is recursively applied to nested groupings of D-Hyperpoints to learn the action discrimination information and creatively integrates Kolmogorov-Arnold Networks (KAN) to enhance spatio-temporal interaction within D-Hyperpoints. Finally, we propose KAN-HyperpointNet, a spatio-temporal decoupled network architecture for 3D action recognition. Extensive experiments on two public datasets: MSR Action3D and NTU-RGB+D 60, demonstrate the state-of-the-art performance of our method.
\end{abstract}

\vspace{-0pt}
\begin{IEEEkeywords}
3D action recognition, point cloud sequence, KAN-HyperpointNet, D-Hyperpoint.
\end{IEEEkeywords}
\end{spacing} 

\vspace{-0.45em}  
\section{Introduction}

With the rapid advancement of 3D sensors (e.g., LiDAR and Kinect), 3D human action recognition has emerged as a critical research focus in computer vision. This task involves identifying action classes from sequences of 3D data, including depth maps, skeleton joints, and point clouds. Particularly, point cloud videos offer distinct advantages for this task due to their capability to provide precise dynamic geometry and shape information, even under complex lighting conditions. However, the irregular and unordered nature of the spatial dimensions in point cloud data presents significant challenges. Traditional grid-based convolutional neural networks (CNNs) are not well-suited to model these data directly, making it difficult to accurately capture and recognize the complex spatio-temporal dynamics inherent in point cloud sequences.

One approach to point cloud sequence modeling is based on voxelization, which converts sequences into regular voxel structures for feature extraction \cite{FaF,MinkowskiNet,3DV}. However, voxelization faces challenges in dynamic 4D sequences due to inefficiencies in handling sparse data and the introduction of quantization errors, leading to computational waste and reduced accuracy, particularly in real-time and high-precision scenarios.

Another research line is focused on directly modeling point cloud sequences by iteratively applying spatio-temporal local encoding to capture motion discriminative features from the microscopic to macroscopic scale \cite{pstnet,pstnet++,p4transformer,psttransformer,mamba4d,3dinaction}. However, due to the spatial irregular properties, iteratively performing spatio-temporal local encoding for point cloud sequences is time-consuming and inhibits parallel computation. Furthermore, spatio-temporal local encoding can result in mutual interference between spatial and temporal information processing, thus compromising the integrity of spatial structure.

To resolve the above issues, several attempts to encode the temporal evolution of static appearances instead of capturing spatio-temporal local structures for recognizing human actions have been proposed \cite{hyperpointnet,squentialpointnet,squentialpointnet2}. In this fashion, space learning and time learning are decoupled, minimizing the impact of the temporal dynamics on spatial structures and enhancing structure integrity of space postures. However, encoding the temporally varying spatial appearance focuses mainly on the macroscopic movements of human actions while ignoring the microscopic dynamics of the limbs, which are crucial cues for inferring complex actions.

\begin{figure}[t]
	\centering
	\includegraphics[width=1.0\columnwidth]{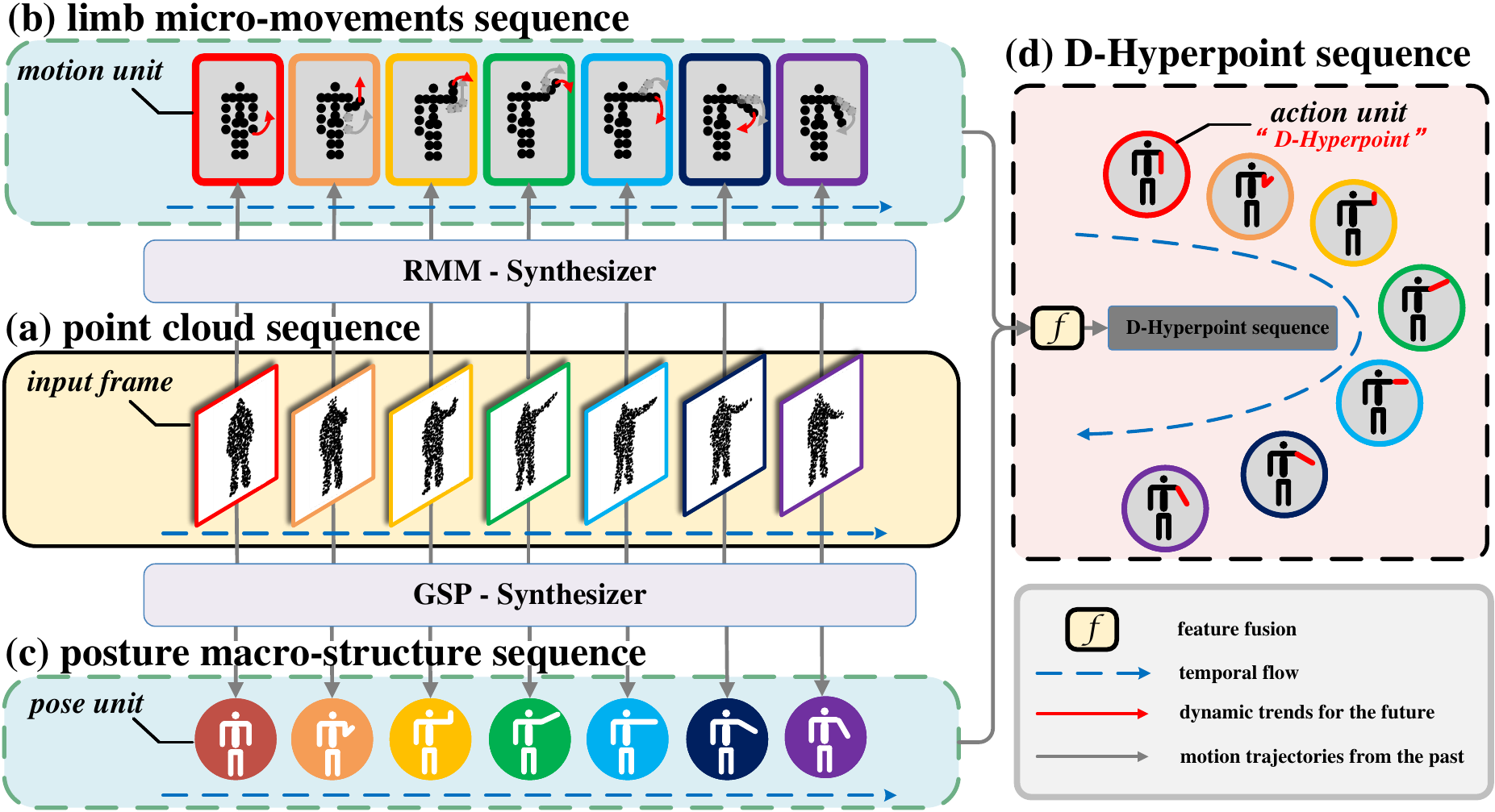} 
	\caption{\fontsize{9pt}{9.5pt}\selectfont Illustration of D-Hyperpoint Sequence Construction. The point cloud sequence (a) is first processed into limb micro-movements sequence (b) and posture macro-structure sequence (c), which are then integrated into a D-Hyperpoint sequence (d) to encapsulate both regional-momentary motion and global-static posture.}
	\label{fig1}
	\vspace{-16.5pt}
\end{figure}

\begin{figure*}[t]
	\centering
	\includegraphics[width=1.0\textwidth, clip=true, trim=0 0 0 0]{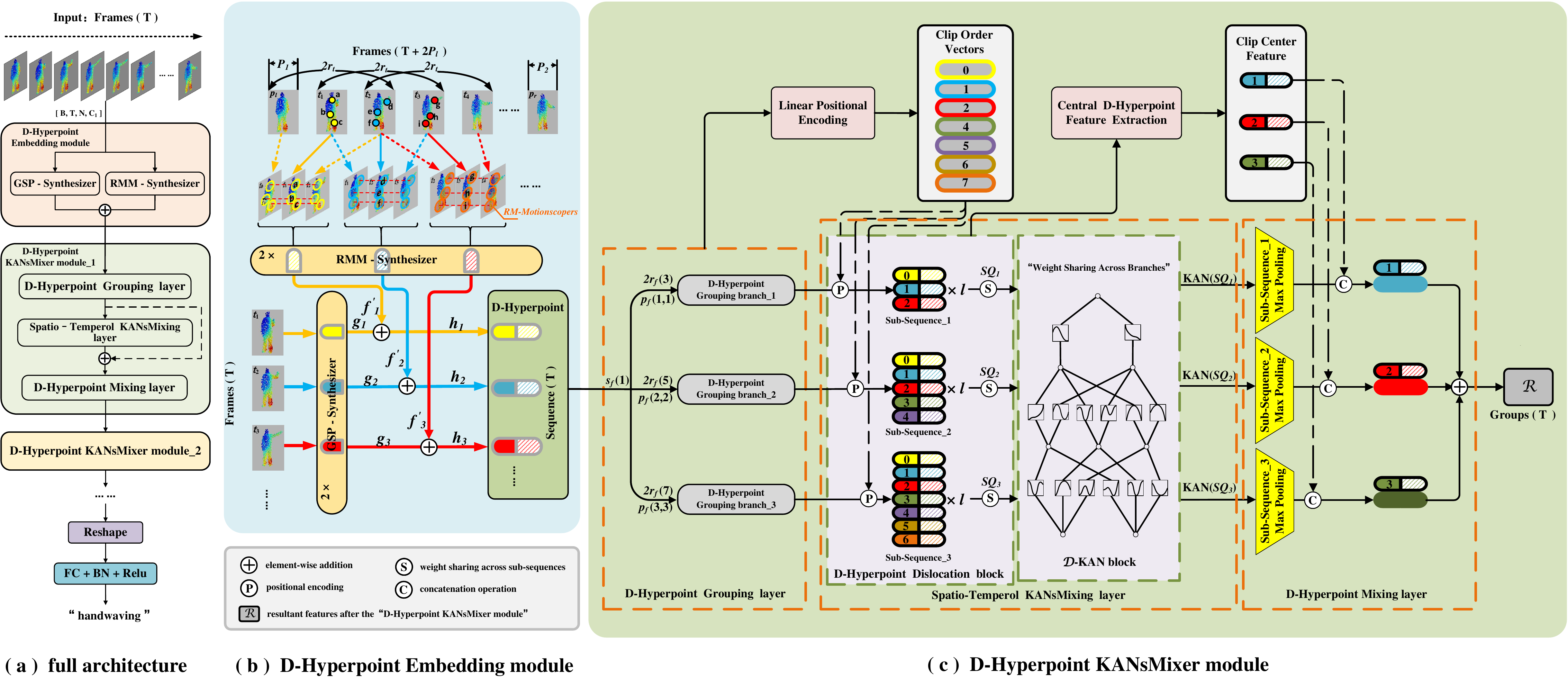}
	\caption{\fontsize{9.5pt}{11pt}\selectfont The overview of KAN-HyperpointNet (a). It contains two modules: D-Hyperpoint Embedding Module (b) and D-Hyperpoint KANsMixer Module (c).}
	\label{fig2}
	\vspace{-15.5pt}
\end{figure*}

In this paper, we propose a novel point cloud sequence network named KAN-HyperpointNet, which takes into account both fine-grained limb micro-movements and integral posture macro-structure to facilitate the accuracy of 3D action recognition. KAN-HyperpointNet consists of two main components: a D-Hyperpoint Embedding module and a D-Hyperpoint KANsMixer module. We first introduce a compact data type, D-Hyperpoint, generated through the D-Hyperpoint Embedding module. This data type integrates the features of regional-momentary limb motion and global-static posture structure, effectively summarizing unit human action at each moment, as shown in Fig.~\ref{fig1}\phantom{.}. Subsequently, we design the D-Hyperpoint KANsMixer module as the fundamental building block, recursively applied to nested groupings of D-Hyperpoints. This module integrates Kolmogorov-Arnold Networks (KAN) \cite{KAN} to enhance the spatio-temporal information interaction of D-Hyperpoints.

Our main contributions are summarized as follows:

\begin{itemize}
	\item We devise a D-Hyperpoint Embedding module to generate a novel data type, termed D-Hyperpoints, for describing sophisticated spatio-temporal actions, which reconciles  the precision of limb micro-movements with the integrity of posture macro-structure.
	\item To the best of our knowledge, we are the first to integrate KAN into a 4D backbone architecture for understanding point cloud sequences. We design a D-Hyperpoint KANsMixer as the fundamental building block for D-Hyperpoint sequence modelling, in which the spatio-temporal interaction of D-Hyperpoints is facilitated by exploiting the superior performance of KAN.
	\item We integrate the D-Hyperpoint Embedding module and the D-Hyperpoint KANsMixer module to design a novel point cloud sequence network called KAN-HyperpointNet. Extensive experiments on NTU RGB+D 60 \cite{ntu60} and MSR-Action3D \cite{msr3daction} datasets demonstrate that KAN-HyperpointNet achieves state-of-the-art performance in point cloud sequence-based 3D human action recognition. Additionally, it provides notable memory and computational efficiency gains.
\end{itemize}

\section{Methodology}
In this section, we present KAN-HyperpointNet, a lightweight and effective point cloud sequence network designed for human action recognition. As illustrated in Fig.~\hyperref[fig2]{\ref*{fig2}(a)}\phantom{.}, the primary components of KAN-HyperpointNet are a D-Hyperpoint Embedding module, a D-Hyperpoint KANsMixer module, and a classifier head.

\subsection{D-Hyperpoint Embedding module}\label{AA}
The D-Hyperpoint Embedding module, which integrates the Regional-Momentary Motion Synthesizer (RMM-Synthesizer) block and the Global-Static Posture Synthesizer (GSP-Synthesizer) block as shown in Fig.~\hyperref[fig2]{\ref*{fig2}(b)}\phantom{.}, is designed to produce a novel data type named D-Hyperpoint. This data type encapsulates both regional-momentary motion and global-static posture of human actions. By organizing D-Hyperpoints into an ordered sequence, we obtain sequential data that leads to superior performance in downstream tasks.

Given a point cloud video of T frames $S=\left\{S_t\right\}_{t=1}^T$, where $S_{t}=\{x_{1}^{t},x_{2}^{t},...,x_{N}^{t}\}$ denotes the $t$-th point cloud frame and $N$  is the number of points. For each frame $s_{t}$ , we use the $L_{t}=\{l_{1}^{t},l_{2}^{t},...,l_{N}^{t}\}\in\mathbb{R}^{N\times3}$ and $F_{t}=\{f_{1}^{t},f_{2}^{t},...,f_{N}^{t}\}\in\mathbb{R}^{N\times C}$ to represent the point cloud coordinates and features.

\textbf{Regional-Momentary Motion Synthesizer block.} The Regional-Momentary Motion Synthesizer (RMM-Synthesizer) block is tailored to abstract the instantaneous motion details of the human limbs in localized areas between adjacent frames, ensuring that subtle changes in complex human movements are effectively captured and represented. Specifically, we apply a temporal stride $d_{t}$ to divide the point cloud video $S$ into $\frac T{d_t}$ equal-length clips, where the center frame of each clip serves as the anchor frame. Every anchor frame has a temporal kernel radius $r_{t}$  around it, encapsulating both the historical motion trajectories and the future movement trends of the limbs. 

Subsequently, Farthest Point Sampling (FPS) \cite{pointnet++} is used to sample anchor points, which are then propagated to neighboring frames in the same clip. Within a spatial radius $r_{s}$ around each anchor point, K nearest neighbor points (KNN) \cite{pointnet++} are selected to construct the Regional-Momentary Motionscoper (RM-Motionscoper), which encapsulates the spatio-temporal local areas around the anchor point.

Once RM-Motionscopers are established, 4D point convolutions are applied to capture localized spatio-temporal features by integrating positional encoding derived from the spatial dimensions $x ,y ,z$ and the temporal dimension  $t$. The convolution operation is defined as:
\vspace{-2pt} 
\begin{equation}
\begin{aligned}
	&f_{ t+\delta_{t}}^{\prime(x+\delta_{x},y+\delta_{y},z+\delta_{z})} \\
	&=\sum_{(\delta_{x},\delta_{y},\delta_{z},\delta_{t})\in C}\gamma(\delta_{x},\delta_{y},\delta_{z},\delta_{t}) \cdot f_{t+\delta_{t}}^{(x+\delta_{x},y+\delta_{y},z+\delta_{z})} \\
	&=\sum_{\delta_t=-r_t}^{r_t}\sum_{\|\delta_x,\delta_y,\delta_z\|\leq r_s}\gamma(\delta_x,\delta_y,\delta_z,\delta_t) \cdot f_{t+\delta_t}^{(x+\delta_x,y+\delta_y,z+\delta_z)} \\
	&=\sum_{||\delta_x,\delta_y,\delta_z||\leq r_z}\mathbf{S}^{(\delta_x,\delta_y,\delta_z)}\cdot\sum_{\delta_t=r_t}^{r_t}\mathbf{T}^{(\delta_t)}\cdot f_{t+\delta_t}^{(x+\delta_x,y+\delta_y,z+\delta_z)}
\end{aligned}
\end{equation}
\vspace{-7.5pt} 

\noindent
where $(\delta_x,\delta_y,\delta_z,\delta_t)$  represents the spatio-temporal displacements, and $\gamma(\delta_x,\delta_y,\delta_z,\delta_t)$  is the convolution kernel function that captures the local body dynamics in our designed RM-Motionscoper $C$. $\mathbf{S}^{(\delta_x,\delta_y,\delta_z)}$ indicates intra-frame aggregation of local instantaneous information, while $\mathbf{T}^{(\delta_t)}$ denotes temporal max pooling for limb dynamic changes. 

In this block, we invoke the RM-Motionscoper twice. During the second application, a unique anchor point is selected to aggregate the limb micro-movement information contained within the entire RM-Motionscoper. This process ultimately flattens the spatio-temporal local features of each clip as $f_t^{\prime}$. Note that to ensure the instantaneous motion information extracted by the RMM-Synthesizer block is injected into each motion unit, frame padding $(p_1,p_2)$ is applied so that the final output frame count matches that of the GSP-Synthesizer block.

\textbf{Global-Static Posture Synthesizer block.} Taking each point cloud frame as input, the Global-Static Posture Synthesizer (GSP-Synthesizer) block encodes each point cloud frame to summarize the posture macro-structure of human, unaffected by time dimension. In this paper, we employ an adaptive attention-enhanced hierarchical spatial encoding architecture, based on PointNet++ \cite{pointnet++}, as the GSP-Synthesizer block to integrate global-static postures. 

The GSP-Synthesizer block comprises three enhanced spatial encoding layers, each formalized as follows:

\vspace{-9.5pt} 
\begin{equation}
	\scalebox{0.98}{$
		\begin{aligned}
			r_j^t=\left[\operatorname*{MAX}_{i=1,\ldots,k_m}\left\{\mathcal{T}\operatorname{-MLP}\left(\left[\left(l_{j,i}^t-o_j^t\right);e_{j,i}^t;f_{j,i}^t\right]\odot A\right)\right\};o_j^t\right]
		\end{aligned}
		$}
\end{equation}
\vspace{-7pt} 

\noindent
where $l_{j,i}^{t}$ is the coordinates of the  $i$-th point in the $j$-th local region of the human body. $o_j^t$ and $f_{j,i}^t$ are the coordinates of the centroid point and the point features corresponding to $l_{j,i}^{t}$, respectively. The Euclidean distance between $l_{j,i}^{t}$ and $o_j^t$ is denoted by $e_{j,i}^t$. $A$ is the attention scores obtained from the Convolutional Block Attention Module \cite{CBAM}. $\odot$ indicates dot product operation. $r_j^t$ represents the abstract features of the $j$-th local region in the $t$-th point cloud frame.

Finally, the D-Hyperpoint integrates the features from both the RMM-Synthesizer and GSP-Synthesizer blocks for each frame .

\subsection{D-Hyperpoint KANsMixer module}\label{AA}
To capture the dynamic information of D-Hyperpoint sequences,
we propose the D-Hyperpoint KANsMixer architecture, which adopts
a temporal hierarchical structure based on the KAN network. Our architecture consists of three main layers: a D-Hyperpoint Grouping layer, a Spatio-Temperol KansMixing (STKM) layer, and a D-Hyperpoint Mixing layer, as depicted in Fig.~\hyperref[fig2]{\ref*{fig2}(c)}\phantom{.}.

\textbf{D-Hyperpoint Grouping layer.} To comprehensively capture dynamic D-Hyperpoint feature variations across different temporal resolutions, we perform D-Hyperpoint grouping on different time scales. Benefiting from the inherently sequential nature of D-Hyperpoint sequence, D-Hyperpoint groups can be determined easily based on the specific temporal radius $r_{f}$, temporal stride $s_{f}$, and temporal padding $p_{f}$. Each scale branch is divided into 
$l$ groups, as shown in Fig.~\hyperref[fig2]{\ref*{fig2}(c)}\phantom{.}

\textbf{Spatio-Temperol KANsMixing layer.} The Spatio-Temperol KANsMixing(STKM) layer is composed of the D-Hyperpoint Dislocation block and the \( \mathcal{D} \)-KAN block, which is presented to abstract each dislocated D-Hyperpoint by mixing its dimension information.

In the D-Hyperpoint Dislocation block, each D-Hyperpoint group is spatially dislocated to record the temporal order by adding the corresponding temporal order vectors. In this fashion, the new coordinate of the D-Hyperpoint can be viewed as the sum of two vectors, namely the temporal order vector and the spatial structure vector: 
\vspace{-4pt} 
\begin{equation}
\begin{aligned}
x_{t+k}= h_{t+k}\oplus ToV_{t_o,h}= h_{t+k}\oplus\left[\left(\frac{t_o}{t_l-1}\right)-0.5\right]
\end{aligned}
\end{equation}
\vspace{-10pt} 

\noindent
where $t_o$ is the temporal position, $h$ is the dimension position, and $t_l$ is the clip length. The $x_{t+k}$  represents the spatio-temporal structural feature of the $k$-th D-Hyperpoint in the $t$-th T-Hyperpoint group after the space dislocation.

Then, the \( \mathcal{D} \)-KAN block integrates the spatial information of dislocated D-Hyperpoints across different temporal marking regions, maps $\mathbb{R}^{d+m}\to\mathbb{R}^{d+m}$, and shares this information among all dislocated D-Hyperpoints. We introduce Kolmogorov-Arnold Networks (KAN) as the core component of the D-Hyperpoint KANsMixer module. KAN is based on the Kolmogorov-Arnold representation theorem and employs trainable B-spline functions to efficiently capture and represent the complex, multivariate relationships inherent in spatio-temporal human motion patterns within point cloud sequences. Compared to conventional neural network structures, KAN exhibits higher computational efficiency with fewer parameters, thereby enhancing the model's flexibility and effectiveness in capturing complex human action patterns \cite{KAN,kagnnskolmogorovarnoldnetworksmeet,kan20}. One \( \mathcal{D} \)-KAN layer is mathematically represented by the following equation:
\vspace{-4pt} 
\begin{equation}
	\begin{aligned}
f(x_{t+k})=\sum_{q=1}^{2N+1}\Phi_q\left(\sum_{n=1}^N\varphi_{q,n}(x_{t+k}^n)\right)
	\end{aligned}
\end{equation}
\vspace{-8pt} 

\noindent
in this formula, $\Phi_q$ represents transformations of human action features, while the motion activation function $\varphi_{q,n}(x_{t+k}^n)$ is expressed as a combination of the basis function $b(x_{t+k})$ and a spline function $\mathrm{spline}(x_{n+k})$, formulated as:
\vspace{-5pt} 
\begin{equation}
	\begin{aligned}
		\varphi(x_{t+k})=w_bb(x_{t+k})+w_s\mathrm{spline}(x_{t+k})
	\end{aligned}
\end{equation}
\vspace{-14pt} 

Here, $b(x_{t+k})$ is defined as:
\vspace{-7pt} 
\begin{equation}
	\begin{aligned}
		b(x_{t+k})=\mathrm{SILU}(x_{t+k})=\frac{x_{t+k}}{1+\exp^{-x}}
	\end{aligned}
\end{equation}
\vspace{-12pt} 

\noindent
where SiLU denotes the Sigmoid Linear Unit activation function. 

The function $\mathrm{spline}(x_{n+k})$, specifically tailored for capturing the nuanced dynamics of human motion, is parameterized as a linear combination of B-splines, expressed as:
\vspace{-6pt} 
\begin{equation}
	\begin{aligned}
		\mathrm{spline}(x_{t+k})=\sum_ic_iB_i(x_{t+k})
	\end{aligned}
\end{equation}
\vspace{-12pt} 

\noindent
where $c_i$ is trainable, $w_b$ and $w_c$ are also trainable coefficients that allow finer control over the activation function, thereby enhancing the effective fusion of spatio-temporal features. \( \mathcal{D} \)-KAN shares weights across multi-stream sub-sequences. 

The \( \mathcal{D} \)-KAN block can be formulated as:
\vspace{-6pt} 
\begin{equation}
	\begin{aligned}
		D\text{-KAN}(x_{t+k})=(\Phi_3\circ\Phi_2\circ\Phi_1)(x_{t+k})
	\end{aligned}
\end{equation}
\vspace{-15pt} 

\textbf{D-Hyperpoint Mixing layer.} The D-Hyperpoint Mixing layer aggregates all D-Hyperpoints from the $t$-th D-Hyperpoint group across each stream, gathers spatial information from different temporal marking regions, and generates the corresponding new D-Hyperpoint feature $F_{t}^{'}$, as demonstrated by the following formula:
\vspace{-5.5pt} 
\begin{equation}
	\begin{aligned}
		F_{t}^{'}={\mathcal T} \mathrm{-MIX}\Big[ {\mathcal D} \mathrm{-KAN}(x_{t+k})+x_{t+k} \Big]\oplus h_{r_{t}-1}
	\end{aligned}
\end{equation}
\vspace{-14pt} 

\noindent
where $h_{r_{t}-1}$ represents the centroid coordinate used for the new D-Hyperpoint in the D-Hyperpoint group.

\section{Experiments}
\subsection{Datasets}\label{AA}

\textbf{NTU RGB+D 60.} NTU RGB+D 60 contains 56,880 samples which are categorized into 60 actions. These samples are performed by 40 volunteers and captured by three Microsoft Kinect v2 cameras from various views concurrently. There are two experiments settings: cross-subject and cross-view settings \cite{ntu60}.

\textbf{MSR Action3D.} The MSR-Action 3D is a widely used small-scale human action dataset, consisting of 557 depth video samples capturing 20 actions performed by 10 subjects \cite{msr3daction}.

\subsection{Training Details}\label{AA}
All experiments are conducted on a machine equipped with an Intel(R) Xeon(R) Platinum 8255C CPU (12 vCPUs, 2.50GHz) and an Nvidia RTX 3090 GPU (24GB), using Python 3.8 and PyTorch 1.10.2 with CUDA 11.3. Our models are trained for 100 epochs with a batch size of 24.  We adopt the data preprocessing following \cite{squentialpointnet} on two datasets.

\vspace{-11pt} 
\begin{table}[b]
	\centering
	\caption{{\fontsize{9pt}{9.5pt}Accuracies (\%) Of Different Methods On The NTU RGB+D 60 Dataset.}}
	\resizebox{\columnwidth}{!}{
		\begin{tabular}{ccccc}
			\hline\hline
			\fontsize{9}{11}\selectfont 
			\multirow{2}{*}{\textbf{Methods}} & \multirow{2}{*}{\textbf{Input}} & \multicolumn{2}{c}{\raisebox{0pt}[0pt][0pt]{\textbf{Accuracy (\%)}}} \\ \cline{3-4} 
			&                                & \textbf{Cross-subject} & \textbf{Cross-view} \\ \hline\hline

			Debnath et al.(2021) \cite{Debnath}           &\multirow{4}{*}{RGB}                            & 87.2                   & —                \\ 
			Piergiovanni et al.(2021) \cite{Piergiovanni}           &                           & —                   & 93.7                 \\ 
			ViewCLR(2023) \cite{ViewCLR}            &                           & 89.7                   &  94.1                \\ 
			Shah et al.(2023) \cite{Shah}           &                           & 91.4                   & 98.0                \\ \hline

			MVDI(2019) \cite{MVDI}           &\multirow{4}{*}{Depth}                            & 84.6                   & 87.3                \\ 
			ADMD(2019) \cite{ADMD}          &                           & 73.1                   & 81.5                \\ 
			3DFCNN(2020) \cite{3dfcnn}          &                           & 78.1                   & 80.4                \\ 
			Stateful ConvLSTM(2020) \cite{ConvLSTM}          &                           & 80.4                   & 79.9                \\ \hline

			ActCLR(2023) \cite{ActCLR}          &\multirow{5}{*}{Skeleton}                           & 88.2                   & 93.9                \\ 
			SkeAttnCLR(2023) \cite{SkeAttnCLR}           &                           &  89.4                   & 94.5                \\ 
			RVTCLR+(2023) \cite{RVTCLR+}           &                           & 87.5                   & 93.9                \\ 
			HiCLR(2023) \cite{HiCLR}           &                           & 90.4                  & 95.7                \\ 
			Js-SaPR-GCN(2024) \cite{Js-SaPR-GCN}          &                           & 90.1                   & 94.9                \\ 
			BlockGCN(2024) \cite{BlockGCN}            &                           & 90.9                   & 95.4                \\ \hline
			
			PSTNet(2021) \cite{pstnet}  &\multirow{8}{*}{Point}                           & 90.5                  & 96.5                \\ 
			PSTNet++(2021) \cite{pstnet++} &                            & 91.4                   & 96.5                \\ 
			P4Transformer(2021) \cite{p4transformer} &                            & 90.2                   & 96.4                \\ 
			PST-Transformer(2022) \cite{psttransformer}  &        & 91.0                   & 96.4                \\ 
			PST-Transformer + MaST-Pre(2023) \cite{MaST-Pre} &                              & 90.8                   & —                \\ 
			PointCPSC(2023) \cite{PointCPSC} &                              & 88.0                   & —                \\ 
			SequentialPointNet(2023) \cite{squentialpointnet} &                              & 90.3                   & 97.6                \\ 
			\textbf{KAN-HyperpointNet(ours)}           &                              & \textbf{91.6}          & \textbf{98.4}       \\ \hline\hline
			
	\end{tabular}}
	\vspace{-9pt} 
	\label{table1}
\end{table}

\begin{table}[t]
	\centering
	\caption{\fontsize{9pt}{9.5pt} Accuracies (\%) Of Different Methods On The MSR Action3D Datasets.}
	\resizebox{7cm}{!}{
		\begin{tabular}{cc}  
			\hline\hline
			\textbf{Methods}                & \textbf{Accuracy (\%)} \\ \hline\hline

			PSTNet(2021) \cite{pstnet}          & 91.20                  \\ 
			P4Transformer(2021) \cite{p4transformer}     & 90.94                  \\ 
			PSTNet++(2021) \cite{pstnet++}         & 92.68                  \\ 
			PST-Transformer(2022) \cite{psttransformer}  & 93.73                  \\ 
			PPTr+C2P(2023) \cite{PPTr+C2P}   & 94.76                  \\ 
			PointCPSC(2023) \cite{PointCPSC}  & 92.68                  \\ 
			PST-Transformer + MaST-Pre(2023) \cite{MaST-Pre}  & 94.08                  \\ 
			SequentialPointNet(2023) \cite{squentialpointnet}  & 92.64                  \\ 
			MAMBA4D(2024) \cite{mamba4d}          & 93.38                  \\ 
			3DInAction(2024) \cite{3dinaction}       & 92.23                  \\ 
			\textbf{KAN-HyperpointNet(ours)}  & \textbf{95.59}         \\ \hline\hline
	\end{tabular}}
	\label{table2}
	\vspace{-9pt} 
\end{table}

\begin{figure}[t]
	\centering
	\includegraphics[width=0.75\columnwidth]{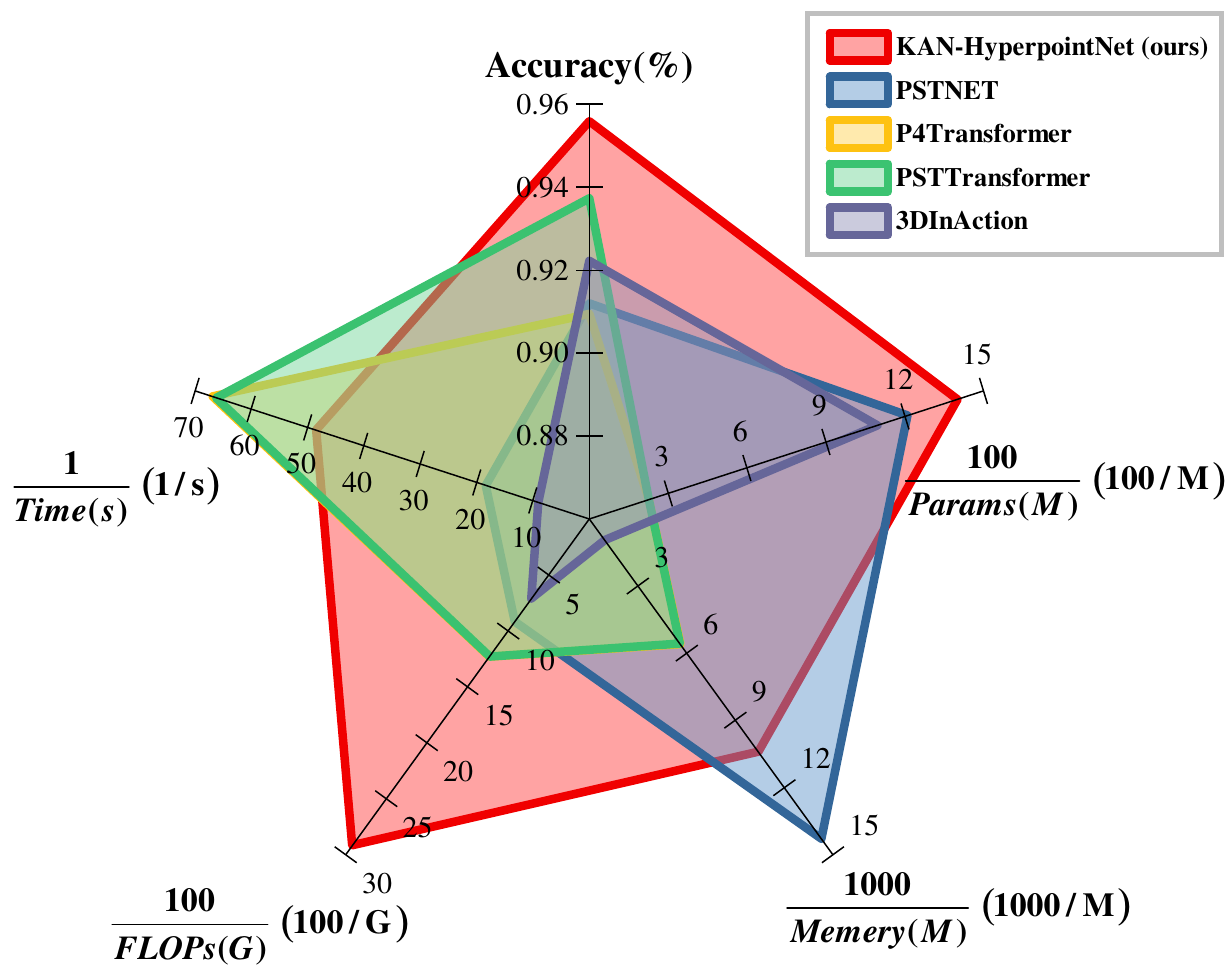} 
	\caption{\fontsize{9pt}{9.5pt}\selectfont Multi-Dimensional Comparison of KAN-HyperpointNet with State-of-the-Art Point Cloud Sequence Networks.}
	\label{fig3}
	\vspace*{-16pt} 
\end{figure}

\vspace*{5pt} 
\subsection{Comparison with the State-of-the-art}\label{AA}

\textbf{Accuracy.} To verify the effectiveness of KAN-HyperpointNet, we conducted comparison experiments on two public datasets. Specifically, on the NTU RGB+D 60 dataset, we compared our model with various advanced depth-based, skeleton-based, and point-based approaches for human action recognition. 

As reported in Table~\ref{table1}\phantom{.}, KAN-HyperpointNet surpasses all state-of-the-art methods in both cross-view and cross-setup settings on the NTU RGB+D 60 dataset. Meanwhile, Table~\ref{table2} reports that KAN-HyperpointNet achieves the highest performance on the MSR Action3D dataset, outperforming the second-best method, PPTr+C2P, by 0.83\%. These results clearly demonstrate the superiority of our method in terms of accuracy.

\textbf{Memory Usage and Computational Efficiency.} We evaluate the memory usage and computational efficiency of KAN-HyperpointNet by comparing its parameter count (Params), maximum running memory (Memory), floating point operation count (FLOPs), and forward inference time (Time) with advanced point cloud sequence networks on the MSR Action3D dataset.

Fig.~\ref{fig3} shows that our method achieves the highest recognition accuracy while significantly reducing modeling complexity and enhancing computational efficiency. These improvements result from KAN-HyperpointNet's time-related parallelism and the efficient edge activations enabled by spline parameterization within KAN \cite{KAN}.

\vspace{-6pt} 
\subsection{Ablation Study}\label{AA}
\vspace{-1pt} 
 In this section, comprehensive ablation studies are performed on the MSR Action3D dataset to validate the contributions of different components in our KAN-HyperpointNet.

\textbf{Effectiveness of D-Hyperpoint Embedding module.} We conduct ablation experiments to compare our proposed D-Hyperpoint with the original Hyperpoint (Baseline) \cite{squentialpointnet} and separately evaluate the performance of the RMM-Synthesizer and GSP-Synthesizer blocks. As illustrated in Table~\ref{table3}\phantom{.}, D-Hyperpoint outperforms both the original Hyperpoint and the individual Synthesizer blocks in terms of accuracy, validating its enhanced ability to preserve the integrity of critical information captured during action inference.

\begin{figure}[t]
	\centering
	\includegraphics[width=0.85\columnwidth]{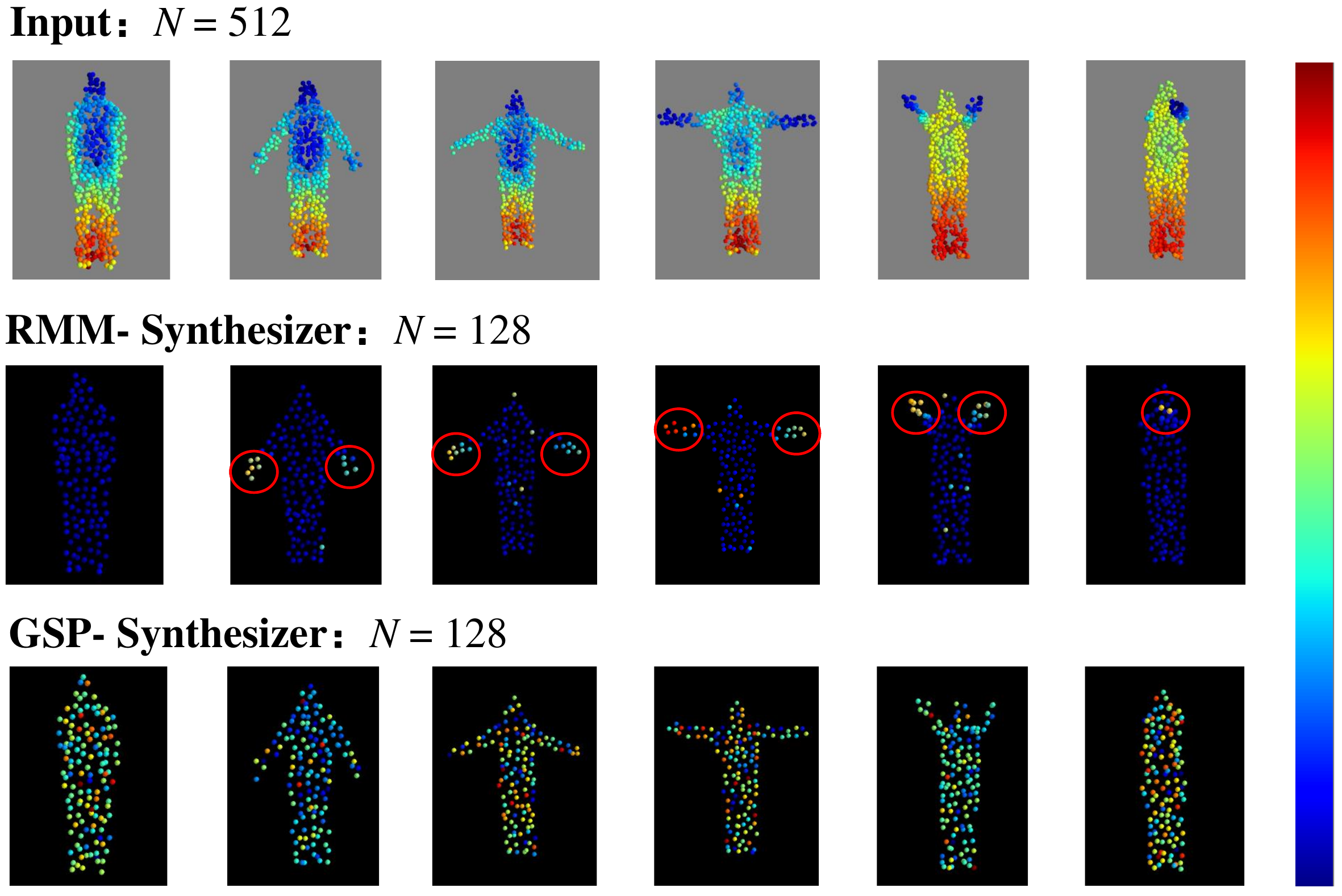} 
	\caption{\fontsize{9pt}{9.5pt}\selectfont Visualization of RMM-Synthesizer and GSP-Synthesizer Outputs in 3D Human Action Recognition.}
	\label{fig4}
	\vspace{-1pt}
\end{figure}

\begin{table}[t]
	\centering
	\caption{\fontsize{9pt}{9.5pt}\selectfont Performance Comparison of Generation and Processing Modules for D-Hyperpoint Sequence.}
	\resizebox{8.5cm}{!}{ 
		\begin{tabular}{cc}  
			\hline\hline
			\textbf{Methods} & \textbf{Accuracy (\%)} \\ \hline\hline
			Baseline\cite{squentialpointnet} & 93.75 \\ 
			KAN-HyperpointNet(w/o RMM-Synthesizer) & 94.12 \\ 
			KAN-HyperpointNet(w/o GSP-Synthesizer) & 89.19 \\ \cline{1-2}
			KAN-HyperpointNet(LSTM) & 82.72 \\ 
			KAN-HyperpointNet(Transformer) & 92.28 \\ 
			KAN-HyperpointNet(Mamba)\cite{mamba} & 91.19 \\ \cline{1-2}
			KAN-HyperpointNet & 95.59 \\ \hline\hline
		\end{tabular}
	}
	\label{table3}
	\vspace{-12pt} 
\end{table}

\textbf{Visualizations.} To investigate what the RMM-Synthesizer block and GSP-Synthesizer block each learn, we visualize their outputs for 3D action recognition in Fig.~\ref{fig4}\phantom{.}. Colors for the input represent depth, while brighter colors indicating higher weights in both synthesizers. In the RMM-Synthesizer, different colors correspond to varying degrees of activation in moving regions, demonstrating the module’s effectiveness in capturing regional-momentary motion information of the human body. In the GSP-Synthesizer block, the color brightness is more uniformly distributed across the overall posture, confirming our hypothesis that the module can efficiently capture global-static posture without temporal influence.

\textbf{Effectiveness of D-Hyperpoint KANsMixer module.} To validate the effectiveness of the D-Hyperpoint KANsMixer module for modeling D-Hyperpoint sequences, we replace it with several well-established time sequence module in our KAN-HyperpointNet.  

As shown in Table~\ref{table3}\phantom{.}, the LSTM and Transformer modules perform much worse when compared with the D-Hyperpoint KANsMixer module. This is because the internal dynamic structures of D-Hyperpoints carry the primary discriminative information, while the changes between D-Hyperpoints are auxiliary, making these modules unsuitable for D-Hyperpoint sequences. Recently, state space models (SSM) like Mamba \cite{mamba} have shown potential for long-sequence modeling in NLP and computer vision \cite{ViM}, but their lack of positional awareness limit their effectiveness in this task.

\section{Conclusion}
In this paper, we introduced KAN-HyperpointNet, a novel network designed to balance fine-grained limb micro-movements and integral posture macro-structure. The D-Hyperpoint Embedding module generates D-Hyperpoints, encapsulating both regional-momentary motion and global-static posture to effectively represent human actions. Additionally, the D-Hyperpoint KANsMixer module enhances spatio-temporal interaction within these D-Hyperpoints using Kolmogorov-Arnold Networks (KAN). Extensive experiments confirm the superior performance of KAN-HyperpointNet in 3D human action recognition.

\clearpage

\renewcommand\refname{\fontsize{10pt}{16pt}\selectfont References}

\fontsize{8.2pt}{9.7pt}\selectfont

\bibliographystyle{IEEEtran}
\bibliography{references}  

\vspace{12pt}

\end{document}